\documentclass[11pt]{article}
\usepackage{emnlp2016}
\usepackage{times}
\usepackage{url}
\usepackage{latexsym}
\usepackage{amsmath}
\usepackage{amssymb}
\usepackage{graphicx}
\usepackage{caption}
\usepackage{subcaption}
\usepackage{alltt}

\emnlpfinalcopy 

\def\todo#1{\bgroup {(#1)}\egroup}

\title{Latent Tree Language Model}

\author{Tom\'{a}\v{s} Brychc\'{i}n \\
NTIS -- New Technologies for the Information Society, \\
Faculty of Applied Sciences, University of West Bohemia, \\
Technick\'{a} 8, 306 14 Plze\v{n}, Czech Republic \\
  {\tt brychcin@kiv.zcu.cz} \\\tt nlp.kiv.zcu.cz \\}

\date{}

\begin{document}
\maketitle
\begin{abstract}

In this paper we introduce Latent Tree Language Model (LTLM), a novel approach to language modeling that encodes syntax and semantics of a given sentence as a tree of word roles. 

The learning phase iteratively updates the trees by moving nodes according to Gibbs sampling. We introduce two algorithms to infer a tree for a given sentence. The first one is based on Gibbs sampling. It is fast, but does not guarantee to find the most probable tree. The second one is based on dynamic programming. It is slower, but guarantees to find the most probable tree. We provide comparison of both algorithms.

We combine LTLM with 4-gram Modified Kneser-Ney language model via linear interpolation. Our experiments with English and Czech corpora show significant perplexity reductions (up to 46\% for English and 49\% for Czech) compared with standalone 4-gram Modified Kneser-Ney language model.

\end{abstract}

\section{Introduction}

Language modeling is one of the core disciplines in natural language processing (NLP). Automatic speech recognition, machine translation, optical character recognition, and other tasks strongly depend on the language model (LM). An improvement in language modeling often leads to better performance of the whole task. The goal of language modeling is to determine the joint probability of a sentence. Currently, the dominant approach is n-gram language modeling, which decomposes the joint probability into the product of conditional probabilities by using the \emph{chain rule}. In traditional n-gram LMs the words are represented as distinct symbols. This leads to an enormous number of word combinations.

In the last years many researchers have tried to capture words contextual meaning and incorporate it into the LMs. Word sequences that have never been seen before receive high probability when they are made of words that are semantically similar to words forming sentences seen in training data. This ability can increase the LM performance because it reduces the \emph{data sparsity} problem. In NLP a very common paradigm for word meaning representation is the use of the \textit{Distributional hypothesis}. It suggests that two words are expected to be semantically similar if they occur in similar contexts (they are similarly distributed in the text) \cite{Harris:1954}. Models based on this assumption are denoted as distributional semantic models (DSMs).

Recently, semantically motivated LMs have begun to surpass the ordinary n-gram LMs.  
The most commonly used architectures are \emph{neural network LMs} \cite{Bengio:2003,Mikolov:2010,Mikolov:2011} and \emph{class-based LMs}. Class-based LMs are more related to this work thus we investigate them deeper.

\newcite{Brown:1992} introduced class-based LMs of English. Their unsupervised algorithm searches classes consisting of words that are most probable in the given context (one word window in both directions). However, the computational complexity of this algorithm is very high. This approach was later extended by \cite{Martin:1998,Whittaker:2003} to improve the complexity and to work with wider context. \newcite{Deschacht2012384} used the same idea and introduced Latent Words Language Model (LWLM), where word classes are latent variables in a graphical model. They apply Gibbs sampling or the expectation maximization algorithm to discover the word classes that are most probable in the context of surrounding word classes. A similar approach was presented in \cite{Brychcin:2014,brychcin:2015}, where the word clusters derived from various semantic spaces were used to improve LMs.

In above mentioned approaches, the meaning of a word is inferred from the surrounding words independently of their relation. An alternative approach is to derive contexts based on the syntactic relations the word participates in. Such syntactic contexts are automatically produced by dependency parse-trees. Resulting word representations are usually less topical and exhibit more functional similarity (they are more syntactically oriented) as shown in \cite{Pado:2007,Levy:2014}.

Dependency-based methods for syntactic parsing have become increasingly popular in NLP in the last years \cite{kubler09}. \newcite{Popel:2010} showed that these methods are promising direction of improving LMs. Recently, unsupervised algorithms for dependency parsing appeared in \cite{Headden:2009,Cohen:2009,Spitkovsky:2010,Spitkovsky:2011,Marecek:2013} offering new possibilities even for poorly-resourced languages. 

In this work we introduce a new DSM that uses tree-based context to create word roles. The word role contains the words that are similarly distributed over similar tree-based contexts. The word role encodes the semantic and syntactic properties of a word. We do not rely on parse trees as a prior knowledge, but we jointly learn the tree structures and word roles.  Our model is a soft clustering, i.e. one word may be present in several roles. Thus it is theoretically able to capture the word polysemy. The learned structure is used as a LM, where each word role is conditioned on its parent role. We present the unsupervised algorithm that discovers the tree structures only from the distribution of words in a training corpus (i.e. no labeled data or external sources of information are needed). In our work we were inspired by class-based LMs \cite{Deschacht2012384}, unsupervised dependency parsing \cite{Marecek:2013}, and tree-based DSMs \cite{Levy:2014}.

This paper is organized as follows. We start with the definition of our model (Section \ref{sec:ltlm}). The process of learning the hidden sentence structures is explained in Section \ref{sec:learning}. We introduce two algorithms for searching the most probable tree for a given sentence (Section \ref{sec:inference}). The experimental results on English and Czech corpora are presented in Section \ref{sec:experiments}. 
We conclude in Section \ref{sec:conclusion} and offer some directions for future work.

\section{Latent Tree Language Model\label{sec:ltlm}}

In this section we describe Latent Tree Language Model (LTLM). LTLM is a generative statistical model that discovers the tree structures hidden in the text corpus.

Let $\boldsymbol L$ be a word vocabulary with total of $|\boldsymbol L|$ distinct words. Assume we have a training corpus $\boldsymbol w$ divided into $S$ sentences. The goal of LTLM or other LMs is to estimate the probability of a text $P(\boldsymbol w)$. Let $N_s$ denote the number of words in the $s$-th sentence. The $s$-th sentence is a sequence of words $\boldsymbol w_s = \{w_{s,i}\}_{i=0}^{N_s}$, where $w_{s,i} \in \boldsymbol L$ is a word at position $i$ in this sentence and $w_{s,0} = \mathrm{<s>}$ is an artificial symbol that is added at the beginning of each sentence.

Each sentence $s$ is associated with the dependency graph $\boldsymbol G_s$. We define the dependency graph as a labeled directed graph, where nodes correspond to the words in the sentence and there is a label for each node that we call \textit{role}. Formally, it is a triple $\boldsymbol G_s=(\boldsymbol V_s,\boldsymbol E_s,\boldsymbol r_s)$ consisting of:

\begin{itemize}
	\item The set of nodes $\boldsymbol V_s = \{0,1,...,N_s\}$. Each token $w_{s,i}$ is associated with node $i \in \boldsymbol V_s$.
	\item The set of edges $\boldsymbol E_s \subseteq \boldsymbol V_s  \times  \boldsymbol V_s$.
	\item The sequence of roles $\boldsymbol r_s = \{r_{s,i}\}_{i=0}^{N_s}$, where $1 \le r_{s,i} \le K$ for $i \in \boldsymbol V_s$.  $K$ is the number of roles. 
\end{itemize}

The artificial word $w_{s,0} = \mathrm{<s>}$ at the beginning of the sentence has always role 1 ($r_{s,0} = 1$). Analogously to $\boldsymbol w$, the sequence of all $\boldsymbol r_s$ is denoted as $\boldsymbol r$ and sequence of all $\boldsymbol G_s$ as $\boldsymbol G$.

Edge $e \in \boldsymbol E_s$ is an ordered pair of nodes $(i,j)$. We say that $i$ is the \textit{head} or the \textit{parent} and $j$ is the \textit{dependent} or the \textit{child}. We use the notation $i \rightarrow j$ for such edge. The directed path from node $i$ to node $j$ is denoted as $i \overset{*}{\rightarrow} j$.

We place a few constraints on the graph $\boldsymbol G_s$.

\begin{itemize}
\item The graph $\boldsymbol G_s$ is a \textit{tree}. It means it is the acyclic graph (if $i \rightarrow j$ then not $j \overset{*}{\rightarrow} i$), where each node has one parent (if $i \rightarrow j$ then not $k \rightarrow j$ for every $k \ne i$).

\item The graph $\boldsymbol G_s$ is \textit{projective} (there are no cross edges). 
For each edge $(i,j)$ and for each $k$ between $i$ and $j$ (i.e. $i < k < j$ or $i > k > j$) there must exist the directed path $i \overset{*}{\rightarrow} k$.

\item The graph $\boldsymbol G_s$ is always rooted in the node 0.
\end{itemize}

\begin{figure}
    \centering
\includegraphics[width=0.49\textwidth]{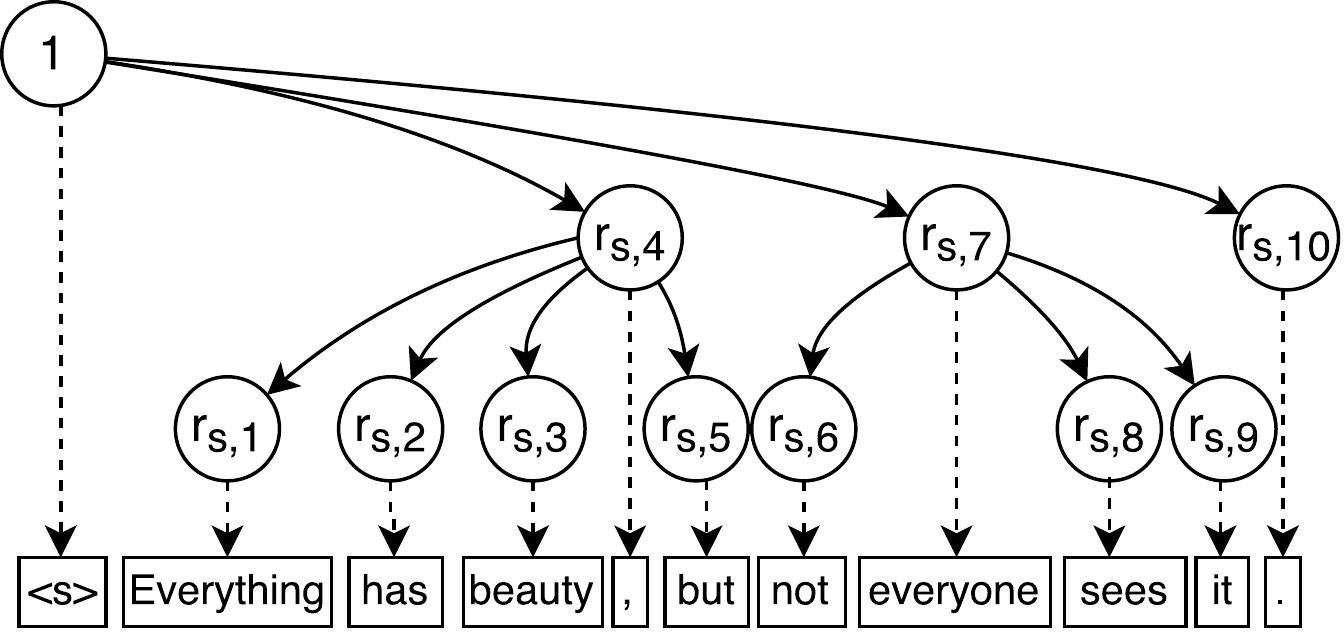}
\caption{\label{fig:tree}Example of LTLM for the sentence "\emph{Everything has beauty, but not everyone sees it.}"}
\end{figure}

We denote these graphs as the \textit{projective dependency trees}. Example of such a tree is on Figure \ref{fig:tree}. 
For the tree $\boldsymbol G_s$ we define a function

\begin{equation}
h_s(j) = i, \quad \text{when }  (i,j) \in \boldsymbol E_s
\end{equation}

\noindent 
that returns the parent for each node except the root.

We use graph $\boldsymbol G_s$ as a representation of the \textit{Bayesian network} with random variables $\boldsymbol E_s$ and $\boldsymbol r_s$. The roles $r_{s,i}$ represent the node labels and the edges express the dependences between the roles. The conditional probability of the role at position $i$ given its parent role is denoted as $P(r_{s,i} |r_{s,{h_s(i)}})$. The conditional probability of the word at position $i$ in the sentence given its role $r_{s,i}$ is denoted as $P(w_{s,i} |r_{s,i} )$.

We model the distribution over words in the sentence $s$ as the mixture

\begin{multline}
P(\boldsymbol w_s) = P(\boldsymbol w_s|r_{s,0}) = \\
\prod\limits_{i = 1}^{N_s} {\sum\limits_{k = 1}^K {P(w_{s,i} |r_{s,i}=k )P(r_{s,i}=k |r_{s,h_s(i)} )} }.
\end{multline}

The root role is kept fixed for each sentence ($r_{s,0}$ = 1) so $P(\boldsymbol w_s) = P(\boldsymbol w_s|r_{s,0})$.

We look at the roles as mixtures over child roles and simultaneously as mixtures over words. We can represent dependency between roles with a set of $K$ multinomial distributions $\boldsymbol \theta$ over $K$ roles, such that $P(r_{s,i}|r_{s,{h_s(i)}} = k) = \theta_{r_{s,i}}^{(k)}$. Simultaneously, dependency of words on their roles can be represented as a set of $K$ multinomial distributions $\boldsymbol \phi$ over $|\boldsymbol L|$ words, such that $P(w_{s,i} |r_{s,i}=k) = \phi_{w_{s,i}}^{(k)}$. To make predictions about new sentences, we need to assume a prior distribution on the parameters $\boldsymbol \theta^{(k)}$ and $\boldsymbol \phi^{(k)}$.

We place a Dirichlet prior $D$ with the vector of $K$ hyper-parameters $\boldsymbol\alpha$ on a multinomial distribution $\boldsymbol \theta^{(k)} \sim D(\boldsymbol\alpha)$ and with the vector of $|\boldsymbol L|$ hyper-parameters $\boldsymbol\beta$ on a multinomial distribution $\boldsymbol \phi^{(k)} \sim D(\boldsymbol\beta)$. In general, $D$ is not restricted to be Dirichlet distribution. It could be any distribution over discrete children, such as logistic normal. In this paper, we focus only on Dirichlet as a conjugate prior to the multinomial distribution and derive the learning algorithm under this assumption.

The choice of the child role depends only on its parent role, i.e. child roles with the same parent are mutually independent. This property is especially important for the learning algorithm (Section \ref{sec:learning}) and also for searching the most probable trees (Section \ref{sec:inference}). We do not place any assumption on the length of the sentence $N_s$ or on how many children the parent node is expected to have.

\section{Parameter Estimation\label{sec:learning}}

In this section we present the learning algorithm for LTLM. The goal is to estimate $\boldsymbol \theta$ and $\boldsymbol \phi$ in a way that maximizes the predictive ability of the model (generates the corpus with maximal joint probability $P(\boldsymbol w)$).

Let $\chi^{k}_{(i,j)}$ be an operation that changes the tree $\boldsymbol G_s$ to $\boldsymbol G_s'$

\begin{equation}
\chi^{k}_{(i,j)}:\boldsymbol G_s \to \boldsymbol G_s',
\end{equation}

\noindent 
such that the newly created tree $\boldsymbol G'(\boldsymbol V_s', \boldsymbol E_s', \boldsymbol r'_s)$ consists of:

\begin{itemize}
\item $\boldsymbol V_s' = \boldsymbol V_s$.
\item $\boldsymbol E_s' = (\boldsymbol E_s \setminus \{(h_s(i),i)\}) \cup \{(j,i)\}$.
\item 
$r'_{s,a} = \left\{\begin{matrix}
r_{s,a} & \textrm{for}~ a \neq i  \\ 
k & \textrm{for}~ a=i
\end{matrix}\right.$,
where $0 \le a \le N_s$.

\end{itemize}

It means that  we change the role of the selected node $i$ so that $r_{s,i}=k$ and simultaneously we change the parent of this node to be $j$. We call this operation a \emph{partial change}.

The newly created graph $\boldsymbol G'$ must satisfy all conditions presented in Section \ref{sec:ltlm}, i.e. it is a projective dependency tree rooted in the node 0. Thus not all partial changes $\chi^{k}_{(i,j)}$ are possible to perform on graph $\boldsymbol G_s$.
 
Clearly, for the sentence $s$ there is at most  $\frac{N_s (1+N_s)}{2}$ parent changes\footnote{The most parent changes are possible for the special case of the tree, where each node $i$ has parent $i-1$. Thus for each node $i$ we can change its parent to any node $j < i$ and keep the projectivity of the tree. That is $\frac{N_s (1+N_s)}{2}$ possibilities.}.

To estimate the parameters of LTLM we apply Gibbs sampling and gradually sample $\chi^{k}_{(i,j)}$ for trees $\boldsymbol G_s$. For doing so we need to determine the posterior predictive distribution\footnote{The posterior predictive distribution is the distribution of an unobserved variable conditioned by the observed data, i.e. $P(X_{n+1}|X_1,...,X_n)$, where $X_i$ are i.i.d. (independent and identically distributed random variables).}

\begin{equation}
\boldsymbol G_s' \sim P(\chi^{k}_{(i,j)}(\boldsymbol G_s)|\boldsymbol w,\boldsymbol G),
\end{equation}

\noindent
from which we will sample partial changes to update the trees. In the equation, $\boldsymbol G$ denote the sequence of all trees for given sentences $\boldsymbol w$ and $\boldsymbol G_s'$ is a result of one sampling. In the following text we derive this equation under assumptions from Section \ref{sec:ltlm}.

The posterior predictive distribution of Dirichlet multinomial has the form of additive smoothing that is well known in the context of language modeling. The hyper-parameters of Dirichlet prior determine how much is the predictive distribution smoothed. Thus the predictive distribution for the word-in-role distribution can be expressed as

\begin{equation}
\label{eq:sample1}
P(w_{s,i} |r_{s,i} ,\boldsymbol w_{\setminus s,i} , \boldsymbol r_{\setminus s,i} ) = \frac{{n_{\setminus {s,i}}^{(w_{s,i}|r_{s,i})} + \beta}}{{n_{\setminus s,i}^{(\bullet| r_{s,i})} + \left|\boldsymbol L \right|\beta }},
\end{equation}

\noindent
where $n_{\setminus {s,i}}^{(w_{s,i}|r_{s,i})}$ is the number of times the role $r_{s,i}$ has been assigned to the word $w_{s,i}$, excluding the position $i$ in the $s$-th sentence. The symbol $\bullet$ represents any word in the vocabulary so that $n_{\setminus s,i}^{(\bullet|r_{s,i})} = \sum_{l \in \boldsymbol L} n_{\setminus {s,i}}^{(l|r_{s,i})}$. We use the symmetric Dirichlet distribution for the word-in-role probabilities as it could be difficult to estimate the vector of hyper-parameters $\boldsymbol \beta$ for large word vocabulary. In the above mentioned equation, $\beta$ is a scalar.

The predictive distribution for the role-by-role distribution is

\begin{equation}
\label{eq:sample2}
P\left( {r_{s,i}| r_{s,h_s(i)},\boldsymbol r_{\setminus s,i} } \right) = \frac{{n_{\setminus {s,i}}^{(r_{s,i}| r_{s,h_s(i)})} + \alpha_{r_{s,i}} }}{{n_{\setminus {s,i}}^{(\bullet|r_{s,h_s(i)})} + \sum\limits_{k = 1}^K {\alpha _k } }}.
\end{equation}

Analogously to the previous equation, $n_{\setminus {s,i}}^{(r_{s,i}|r_{s,h_s(i)})}$ denote the number of times the role $r_{s,i}$ has the parent role $r_{s,h_s(i)}$, excluding the position $i$ in the $s$-th sentence. The symbol $\bullet$ represents any possible role to make the probability distribution summing up to 1. We assume an asymmetric Dirichlet distribution.

We can use predictive distributions of above mentioned Dirichlet multinomials to express the joint probability that the role at position $i$ is $k$ ($r_{s,i} = k$) with parent at position $j$ conditioned on current values of all variables, except those in position $i$ in the sentence $s$

\begin{multline}
\label{eq:joint}
P(r_{s,i}=k,j|\boldsymbol w,\boldsymbol r_{\setminus s,i})  \propto \\
\begin{array}{*{50}l}
   &  P(w_{s,i}|r_{s,i}=k,\boldsymbol w_{\setminus s,i},\boldsymbol r_{\setminus s,i}) \\
   \times &  P(r_{s,i}=k|r_{s,j},\boldsymbol r_{\setminus s,i})  \\
   \times & \!\!\!\!\!\! \prod\limits_{a:h_s(a)=i} \!\!\!\!\!\! P(r_{s,a}|r_{s,i}=k,\boldsymbol r_{\setminus s,i}). \\
\end{array}
\end{multline}

\noindent
The choice of the node $i$ role affects the word that is produced by this role and also all the child roles of the node $i$. Simultaneously, the role of the node $i$ depends on its parent $j$ role. Formula \ref{eq:joint} is derived from the joint probability of a sentence $s$ and a tree $\boldsymbol G_s$, where all probabilities which do not depend on the choice of the role at position $i$ are removed and equality is replaced by proportionality ($\propto$).

We express the final predictive distribution for sampling partial changes $\chi^{k}_{(i,j)}$ as

\begin{equation}
\label{eq:predictive}
P(\chi^{k}_{(i,j)}(\boldsymbol G_s)|\boldsymbol w,\boldsymbol G) \propto \frac{P(r_{s,i}=k,j|\boldsymbol w,\boldsymbol r_{\setminus s,i})}{P(r_{s,i},h_s(i)|\boldsymbol w,\boldsymbol r_{\setminus s,i})}
\end{equation}

\noindent
that is essentially the fraction between the joint probability of $r_{s,i}$ and its parent after the partial change and before the partial change (conditioned on all other variables). This fraction can be interpreted as the necessity to perform this partial change.

We investigate two strategies of sampling partial changes:

\begin{itemize}
\item \textbf{Per sentence}: We sample a single partial change according to Equation \ref{eq:predictive} for each sentence in the training corpus. It means during one pass through the corpus (one iteration) we perform $S$ partial changes.
\item \textbf{Per position}:  We sample a partial change for each position in each sentence. We perform in total $N = \sum_{s=1}^S N_s$ partial changes during one pass. Note that the denominator in Equation \ref{eq:predictive} is constant for this strategy and can be removed.

\end{itemize}

We compare both training strategies in Section \ref{sec:experiments}. After enough training iterations, we can estimate the conditional probabilities $\phi_{l}^{(k)}$ and $\theta_{k}^{(p)}$ from actual samples as

\begin{equation}
\label{eq:final1}
\phi_{l}^{(k)}  \approx \frac{{n^{(w_{s,i}=l|r_{s,i}=k)} + \beta}}{{n^{(\bullet| r_{s,i}=k)} + \left|\boldsymbol L \right|\beta }}
\end{equation}

\begin{equation}
\label{eq:final2}
\theta_{k}^{(p)}  \approx \frac{{n^{(r_{s,i}=k| r_{s,h_s(i)}=p)} + \alpha_{k} }}{{n^{(\bullet|r_{s,h_s(i)}=p)} + \sum\limits_{m = 1}^K {\alpha _m } }}.
\end{equation}

\noindent
These equations are similar to equations \ref{eq:sample1} and \ref{eq:sample2}, but here the counts $n$ do not exclude any position in a corpus.

Note that in the Gibbs sampling equation, we assume that the Dirichlet parameters $\boldsymbol \alpha$ and $\beta$ are given. We use a fixed point iteration technique described in \cite{Minka03} to estimate them.

\section{Inference\label{sec:inference}}

In this section we present two approaches for searching the most probable tree for a given sentence assuming we have already estimated the parameters $\boldsymbol \theta$ and $\boldsymbol \phi$.

\subsection{Non-deterministic Inference}

We use the same sampling technique as for estimating parameters (Equation \ref{eq:predictive}), i.e. we iteratively sample the partial changes $\chi^{k}_{(i,j)}$. However, we use equations \ref{eq:final1} and \ref{eq:final2} for predictive distributions of Dirichlet multinomials instead of \ref{eq:sample1} and \ref{eq:sample2}. In fact, these equations correspond to the predictive distributions over the newly added word $w_{s,i}$ with the role $r_{s,i}$ into the corpus, conditioned on $\boldsymbol w$ and $\boldsymbol r$. This sampling technique rarely finds the best solution, but often it is very near.

\subsection{Deterministic Inference}

\begin{figure}[Ht!]
     \begin{center}
	\begin{subfigure}{0.45\textwidth}
           	 \includegraphics[width=\textwidth]{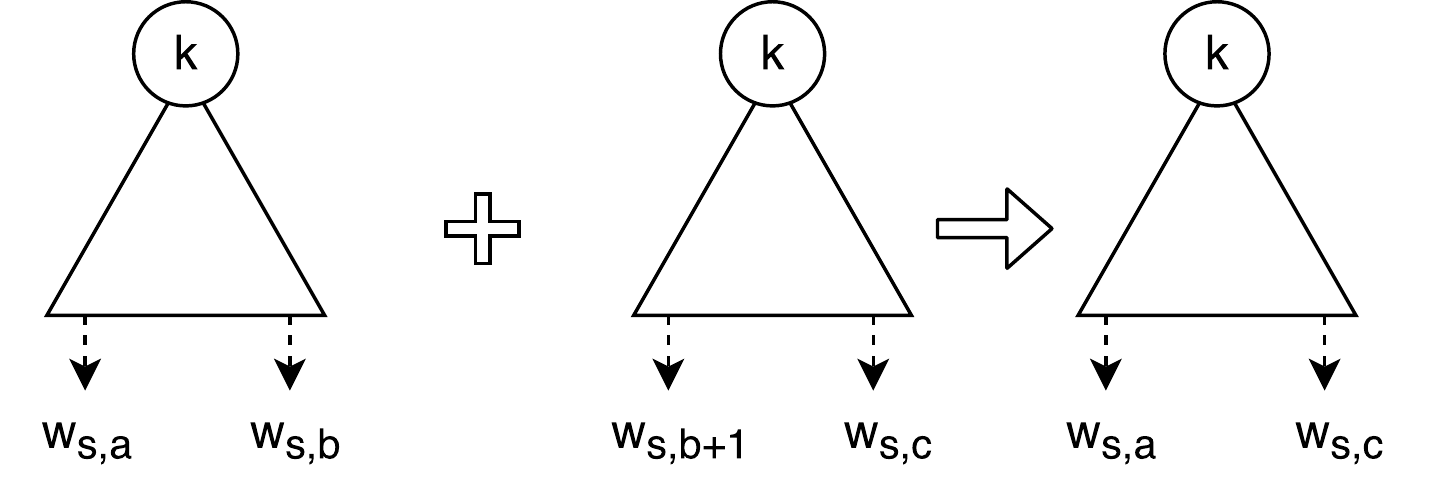}
		\caption{\label{fig:twoandmore}The root has two or more children.}
         \end{subfigure}\qquad
	\begin{subfigure}{0.45\textwidth}
          	\includegraphics[width=\textwidth]{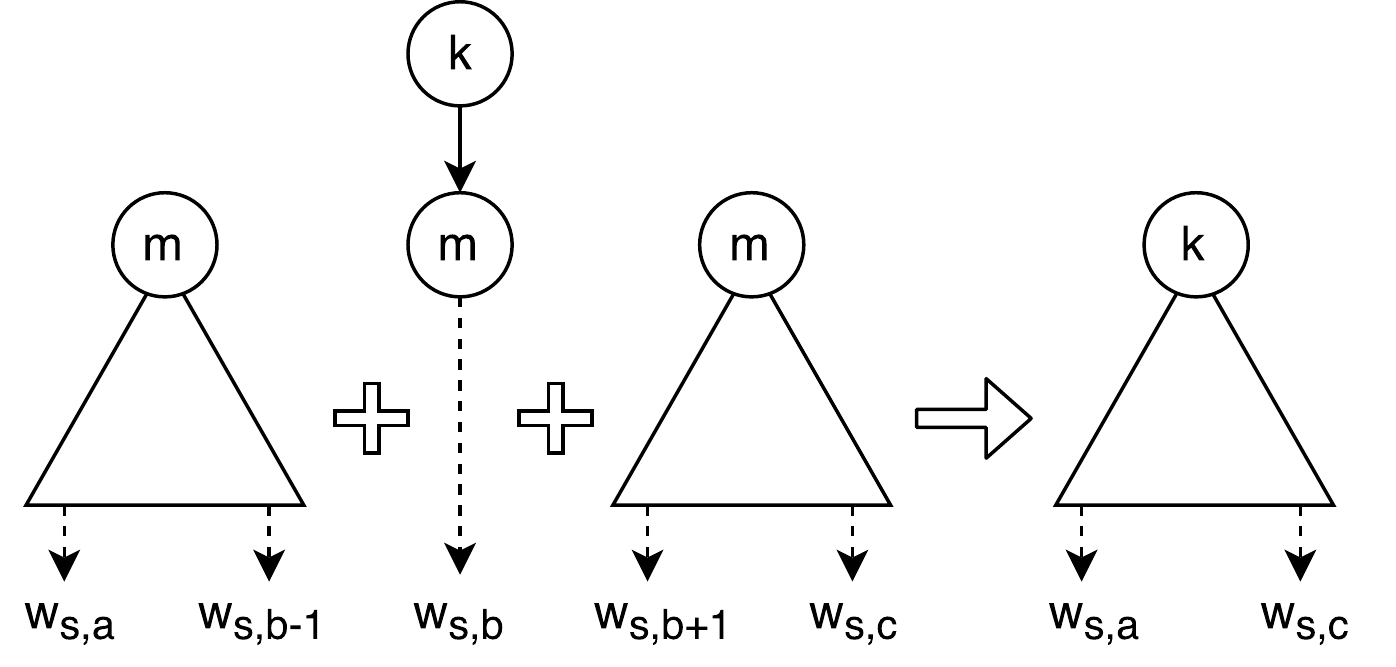}
		\caption{\label{fig:one}The root has only one child.}
        \end{subfigure}
\caption{Searching the most probable subtrees.}
\end{center}
\end{figure}

Here we present the deterministic algorithm that guarantees to find the most probable tree for a given sentence. We were inspired by Cocke-Younger-Kasami (CYK) algorithm \cite{Lange09}.

Let $\boldsymbol T_{s,a,c}^n$ denote the subtree of $\boldsymbol G_s$ (subgraph of $\boldsymbol G_s$ that is also a tree) containing subsequence of nodes $\{a,a+1,...,c\}$. The superscript $n$ denotes the number of children the root of this subtree has. We denote the joint probability of a subtree from position $a$ to position $c$ with the corresponding words conditioned by the root role $k$ as $P^{n}(\{w_{s,i}\}_{i=a}^c,\boldsymbol T^{n}_{s,a,c}|k)$. Our goal is to find the tree $\boldsymbol G_s = \boldsymbol T_{s,0,N_s}^{1+}$ that maximizes probability $P(\boldsymbol w_s, \boldsymbol G_s) = P^{1+}(\{w_{s,i}\}_{i=0}^{N_s},\boldsymbol T_{s,0,N_s}^{1+}|0)$.

Similarly to CYK algorithm, our approach follows bottom-up direction and goes through all possible subsequences for a sentence (sequence of words). At the beginning, the probabilities for subsequences of length 1 (i.e. single words) are calculated as $P^{1+}(\{w_{s,a}\},\boldsymbol T_{s,a,a}^{1+}|k) = P(w_{s,a}|r_{s,a}=k)$. Once it has considered subsequences of length 1, it goes on to subsequences of length 2, and so on.

Thanks to mutual independence of roles under the same parent, we can find the most probable subtree with the root role $k$ and with at least two root children according to

\begin{multline}
P^{2+}(\{w_{s,i}\}_{i=a}^c,\boldsymbol T^{2+}_{s,a,c}|k) = \mathop {\max }\limits_{b:a < b < c} \\
[ P^{1+}(\{w_{s,i}\}_{i=a}^b,\boldsymbol T_{s,a,b}^{1+}|k) \times \\
P^{1+}(\{w_{s,i}\}_{i=b+1}^c,\boldsymbol T_{s,b+1,c}^{1+}|k) ].
\end{multline}

\noindent
It means we merge two neighboring subtrees with the same root role $k$. This is the reason why the new subtree has at least two root children. This formula is visualized on Figure \ref{fig:twoandmore}. Unfortunately, this does not cover all subtree cases. We find the most probable tree with only root child as follows

\begin{multline}
P^{1}(\{w_{s,i}\}_{i=a}^c,\boldsymbol T^{1}_{s,a,c}|k) =  \mathop {\max }\limits_{b,m:a \le b \le c, 1 \le m \le K}\\
 [ P(w_{s,b}|r_{s,b}=m) \times P(r_{s,b}=m|k)  \times \\
P^{1+}(\{w_{s,i}\}_{i=a}^{b-1},\boldsymbol T_{s,a,b-1}^{1+}|m)  \times \\
P^{1+}(\{w_{s,i}\}_{i=b+1}^c,\boldsymbol T_{s,b+1,c}^{1+}|m) ].
\end{multline}

\noindent
This formula is visualized on Figure \ref{fig:one}.

To find the most probable subtree no matter how many children the root has, we need to take the maximum from both mentioned equations $P^{1+} = max(P^{2+}, P^{1})$.

The algorithm has complexity $\mathcal{O}(N_s^3 K^2)$, i.e. it has cubic dependence on the length of the sentence $N_s$.

\section{Side-dependent LTLM}

Until now, we presented LTLM in its simplified version. In role-by-role probabilities (role conditioned on its parent role) we did not distinguish whether the role is on the left side or the right side of the parent. However, this position keeps important information about the syntax of words (and their roles).

We assume separate multinomial distributions $\boldsymbol {\dot\theta}$ for roles that are on the left and $\boldsymbol {\ddot\theta}$ for roles on the right. Each of them has its own Dirichlet prior with hyper-parameters $\boldsymbol {\dot\alpha}$ and $\boldsymbol {\ddot\alpha}$, respectively. The process of estimating LTLM parameters is almost the same. The only difference is that we need to redefine the predictive distribution for the role-by-role distribution (Equation \ref{eq:sample2}) to include only counts of roles on the appropriate side. Also, every time the role-by-role probability is used we need to distinguish sides:

\begin{equation}
P( r_{s,i}| r_{s,h_s(i)})=\left\{\begin{matrix}
\dot\theta_{r_{s,i}}^{(r_{s,h_s(i)})} & \textrm{for}~i < h_s(i))\\ 
\ddot\theta_{r_{s,i}}^{(r_{s,h_s(i)})} & \textrm{for}~i > h_s(i))
\end{matrix}\right..
\end{equation}

In the following text we always assume the side-dependent LTLM.

\section{Experimental Results and Discussion\label{sec:experiments}}

In this section we present experiments with LTLM on two languages, English (EN) and Czech (CS).

As a training corpus we use CzEng 1.0 \cite{czeng10:lrec2012} of the sentence-parallel Czech-English corpus. We choose this corpus because it contains multiple domains, it is of reasonable length, and it is parallel so we can easily provide comparison between both languages. The corpus is divided into 100 similarly-sized sections. We use parts 0--97 for training, the part 98 as a development set, and the last part 99 for testing.

We have removed all sentences longer than 30 words. The reason was that the complexity of the learning phase and the process of searching most probable trees depends on the length of sentences. It has led to removing approximately a quarter of all sentences. The corpus is available in a tokenized form so the only preprocessing step we use is lowercasing. We keep the vocabulary of 100,000 most frequent words in the corpus for both languages. The less frequent words were replaced by the symbol $<$unk$>$. Statistics for the final corpora are shown in Table \ref{tab:corpora}.

\begin{table}[Ht!]
\resizebox{0.49\textwidth}{!}{
\begin{tabular}{lrrr}
\hline
\bf Corpora &  \bf Sentences & \bf Tokens & \bf OOV rate \\
\hline
 EN train & 11,530,604 & 138,034,779 & 1.30\% \\
 EN develop. & 117,735 & 1,407,210  & 1.28\%  \\
 EN test & 117,360 & 1,405,106 & 1.33\% \\
 CS train & 11,832,388 & 133,022,572 & 3.98\% \\
 CS develop. & 120,754 &  1,353,015 & 4.00\% \\
 CS test & 120,573 & 1,357,717 & 4.03\% \\
\hline
\end{tabular}
}
\caption{\label{tab:corpora}Corpora statistics. \emph{OOV rate} denotes the out-of-vocabulary rate.}
\end{table}

\begin{figure}
\includegraphics[width=0.48\textwidth]{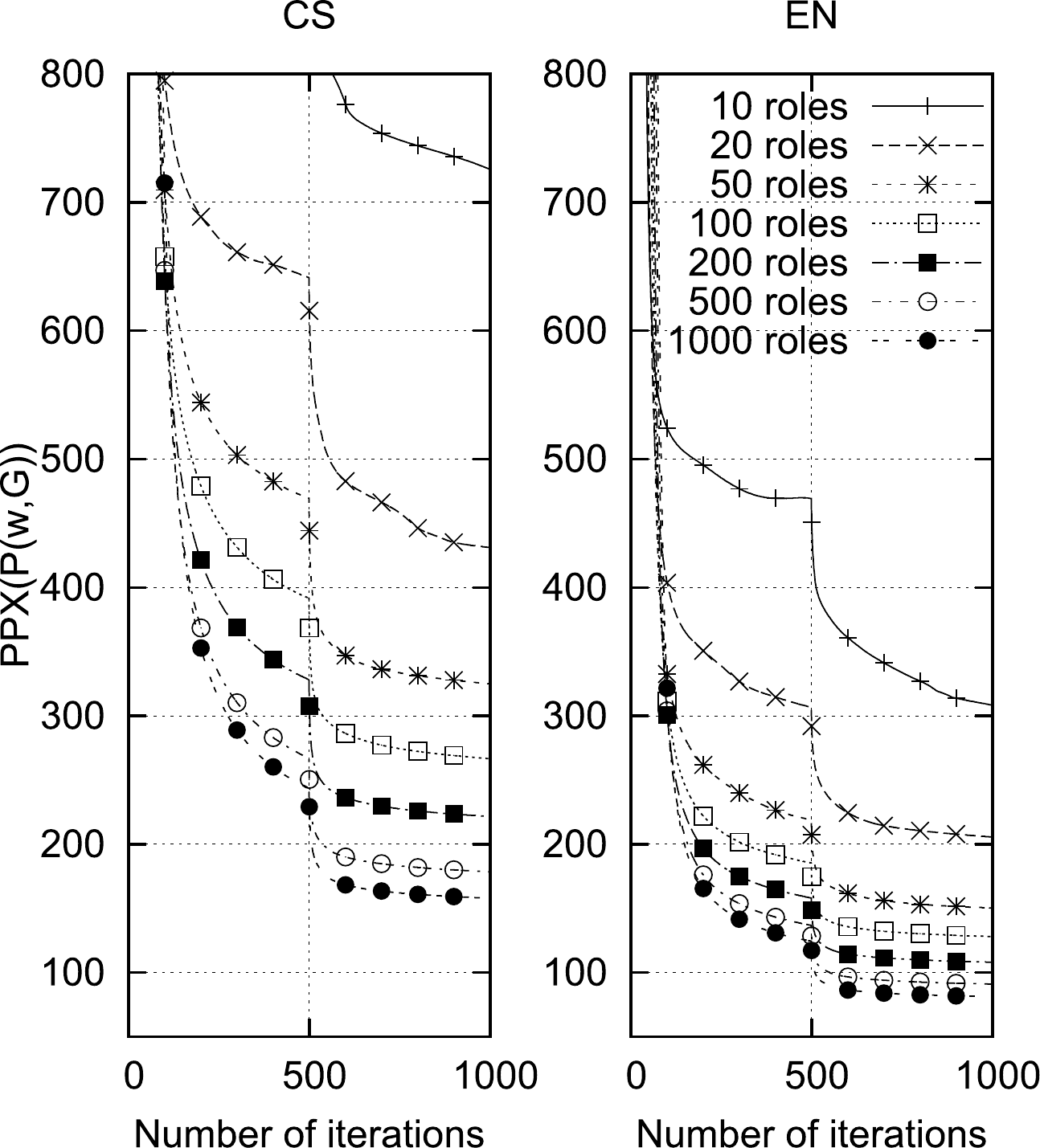}
\caption{\label{fig:learning}Learning curves of LTLM for both English and Czech. The points in the graphs represent the perplexities in every 100th iteration. }
\end{figure}

We measure the quality of LTLM by \emph{perplexity} that is the standard measure used for LMs. Perplexity is a measure of uncertainty. The lower perplexity means the better predictive ability of the LM.

During the process of parameter estimation we measure the perplexity of joint probability of sentences and their trees defined as $\mathrm{PPX}(P(\boldsymbol w,\boldsymbol G)) = \sqrt[N]{{\frac{1}{{P(\boldsymbol w,\boldsymbol G)}}}}$, where $N$ is the number of all words in the training data $\boldsymbol w$.

As we describe in Section \ref{sec:learning}, there are two approaches for the parameter estimation of LTLM. During our experiments, we found that the per-position strategy of training has the ability to converge faster, but to a worse solution compared to the per-sentence strategy which converges slower, but to a better solution. 

We train LTLM by 500 iterations of the per-position sampling followed by another 500 iterations of the per-sentence sampling. This proves to be efficient in both aspects, the reasonable speed of convergence and the satisfactory predictive ability of  the model. The learning curves are showed on Figure \ref{fig:learning}. We present the models with 10, 20, 50, 100, 200, 500, and 1000 roles. The higher role cardinality models were not possible to create because of the very high computational requirements. Similarly to the training of LTLM, the non-deterministic inference uses 100 iterations of per-position sampling followed by 100 iterations of per-sentence sampling.

In the following experiments we measure how well LTLM generalizes the learned patterns, i.e. how well it works on the previously unseen data. Again, we measure the perplexity, but of probability $P(\boldsymbol w)$ for mutual comparison with different LMs that are based on different architectures ($\mathrm{PPX}(P(\boldsymbol w)) = \sqrt[N]{{\frac{1}{{P(\boldsymbol w)}}}}$). 

\begin{table}[Ht!]
\centering
\setlength{\tabcolsep}{0.3em}
\resizebox{0.49\textwidth}{!}{
\begin{tabular}{rrr}
\hline
\bf Model  & \bf EN & \bf CS  \\
\hline
2-gram MKN	& 165.9 & 272.0 \\
3-gram MKN	& 67.7 & 99.3 \\
4-gram MKN	& \bf 46.2 & 73.5 \\
300n RNNLM	& 51.2 & \bf 69.4 \\
4-gram LWLM & 52.7 & 81.5 \\
PoS STLM	& 455.7  & 747.3 \\
1000r STLM  & 113.7 & 211.0 \\
1000r det. LTLM & 54.2 & 111.1 \\
\hline
4-gram MKN + 300n RNNLM	& 36.8 (-20.4\%) & 49.5 (-32.7\%)  \\
4-gram MKN + 4-gram LWLM & 41.5 (-10.2\%) & 62.4 (-15.1\%) \\
4-gram MKN + PoS STLM	& 42.9 (-7.1\%) & 63.3 (-13.9\%) \\
4-gram MKN + 1000r STLM & 33.6 (-27.3\%) & 50.1 (-31.8\%) \\
4-gram MKN + 1000r det. LTLM & \bf 24.9 (-43.1\%) & \bf 37.2 (-49.4\%) \\
\hline
\end{tabular}
}
\caption{\label{tab:baseline} Perplexity results on the test data. The numbers in brackets are the relative improvements compared with standalone 4-gram MKN LM.}
\end{table}

\begin{table*}[Ht!]
\begin{center}
\setlength{\tabcolsep}{0.3em}
\resizebox{\textwidth}{!}{
\begin{tabular}{r|rrrrrrr|rrrrrrr}
\hline
& \multicolumn{7}{c}{\bf EN} & \multicolumn{7}{|c}{\bf CS}  \\
\bf Model\textbackslash roles  & \bf10 & \bf20 & \bf50 & \bf100 & \bf200 & \bf500 & \bf1000 & \bf10 & \bf20 & \bf50 & \bf100 & \bf200 & \bf500 & \bf1000    \\
\hline
STLM & 408.5 & 335.2 & 261.7 & 212.6 & 178.9 & 137.8 & 113.7 & 992.7 & 764.2 & 556.4 & 451.0 & 365.9 & 265.7 & 211.0 \\
non-det. LTLM & 329.5 &	215.1 &	160.4 &	126.5 &	105.6 &	86.7 &	78.4 	& 851.0 &	536.6 &	367.4 &	292.6 &	235.2 &	186.1 &	157.6 \\
det. LTLM & 252.4 & 	166.4 & 	115.3 & 	92.0 & 	75.4 & 	60.9 & 	54.2 	& 708.5 &	390.2 &	267.8 &	213.2 &	167.9 &	133.5 &	111.1 \\
\hline
4-gram MKN + STLM & 42.7 & 41.6 & 39.9 & 37.9 &	 36.3 & 34.9 & 33.6 & 67.5 & 65.1 & 61.4 & 58.3 & 55.5 & 52.4 & 50.1 \\
4-gram MKN + non-det. LTLM & 41.1 &	38.0 &	35.2 &	32.7 &	30.7 &	28.9 &	27.8 	& 65.8 &	59.4 &	55.1 &	51.1 &	47.5	 &43.7 &	41.3 \\
4-gram MKN + det. LTLM & 39.9 &	 36.4 &	32.8 &	30.3 &	28.1 &	26.0 &	\bf 24.9 	& 64.4 & 	56.1 & 	51.5 & 	47.3 &	43.4 &	39.9 &	\bf 37.2\\
\hline
\end{tabular}
}
\end{center}
\caption{\label{tab:results}Perplexity results on the test data for LTLMs and STLMs with different number of roles. Deterministic inference is denoted as \emph{det.} and non-deterministic inference as \emph{non-det.}}
\end{table*}

To show the strengths of LTLM we compare it with several state-of-the-art LMs. We experiment with Modified Kneser-Ney (MKN) interpolation \cite{Chen:1998}, with Recurrent Neural Network LM (RNNLM) \cite{Mikolov:2010,Mikolov:2011}\footnote{Implementation is available at \url{http://rnnlm.org/}. Size of the hidden layer was set to 300 in our experiments. It was computationally intractable to use more neurons.}, and with LWLM \cite{Deschacht2012384}\footnote{Implementation is available at \url{http://liir.cs.kuleuven.be/software.php}.}. We have also created syntactic dependency tree based LM (denoted as STLM). Syntactic dependency trees for both languages are provided within CzEng corpus and are based on MST parser \cite{McDonald:2005}. We use the same architecture as for LTLM and experiment with two approaches to represent the roles. Firstly, the roles are given by the part-of-speech tag (denoted as PoS STLM). No training is required, all information come from CzEng corpus. Secondly, we learn the roles using the same algorithm as for LTLM. The only difference is that the trees are kept unchanged. Note that both deterministic and non-deterministic inference perform almost the same in this model so we do not distinguish between them. 

We combine baseline 4-gram MKN model with other models via linear combination (in the tables denoted by the symbol $+$) that is simple but very efficient technique to combine LMs. Final probability is then expressed as

\begin{equation}
P(\boldsymbol w) = \prod\limits_{s = 1}^S {\prod\limits_{i = 1}^{N_s } {\left[ {\lambda P^{\textrm{LM1}}  + \left( {\lambda  - 1} \right)P^{\textrm{LM2}} } \right]} }.
\end{equation}

\noindent
In the case of MKN the probability $P^{\textrm{MKN}}$ is the probability of a word $w_{s,i}$ conditioned by 3 previous words with MKN smoothing. For LTLM or STLM this probability is defined as

\begin{multline}
P^{\textrm{LTLM}} (w_{s,i} |r_{s,h_s (i)} ) =\\
 \sum\limits_{k = 1}^K {P(w_{s,i} |r_{s,i}  = k)} P(r_{s,i}  = k|r_{s,h_s (i)} ).
\end{multline}

\begin{figure}[Ht!]
    \centering
\includegraphics[width=0.48\textwidth]{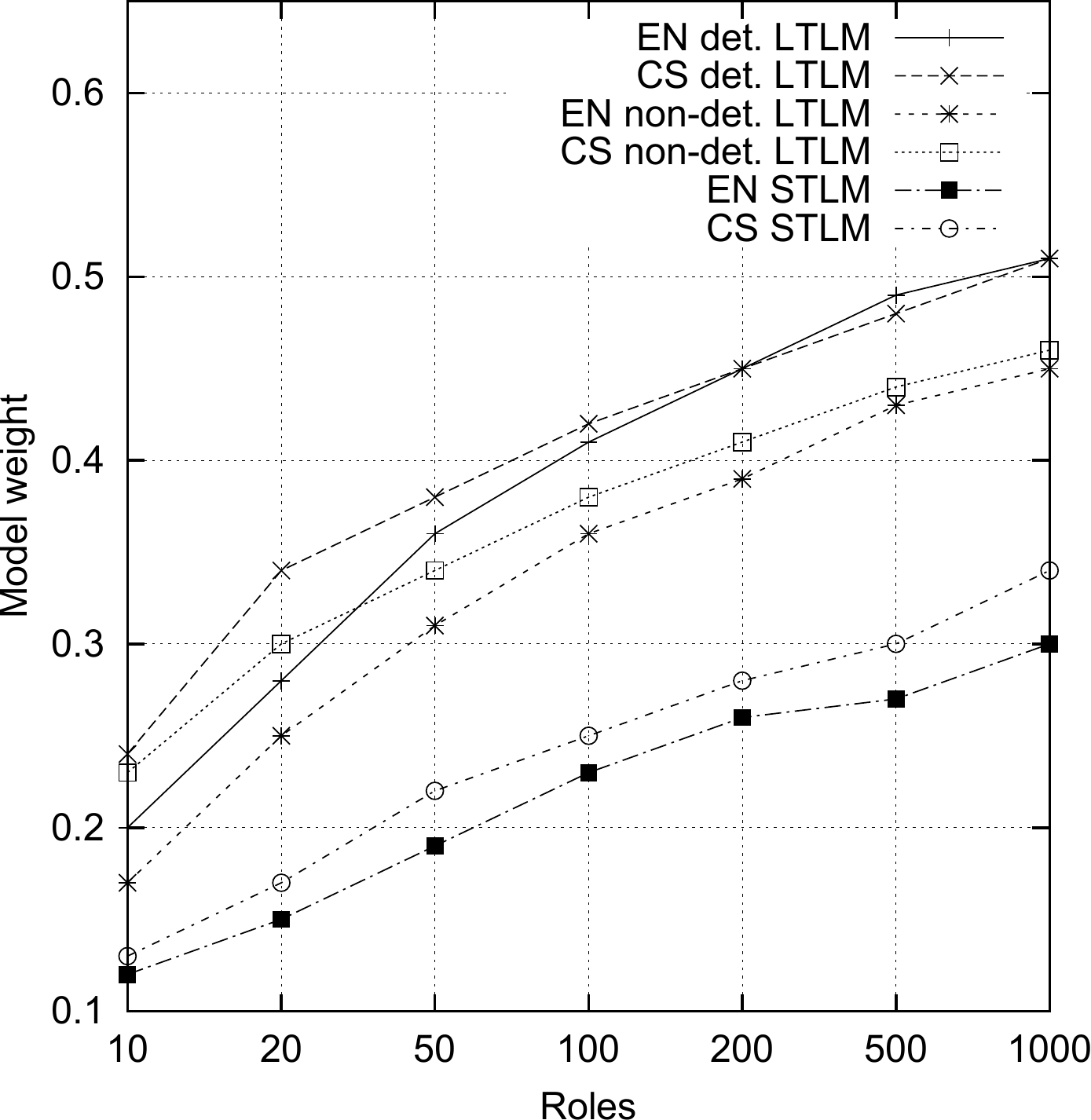}
\caption{\label{fig:weights}Model weights optimized on development data when interpolated with 4-gram MKN LM.}
\end{figure}

We use the \textit{expectation maximization} algorithm \cite{Dem:1977} for the maximum likelihood estimate of $\lambda$ parameter on the development part of the corpus. The influence of the number of roles on the perplexity is shown in Table \ref{tab:results} and the final results are shown in Table \ref{tab:baseline}. Note that these perplexities are not comparable with those on Figure \ref{fig:learning}  ($\mathrm{PPX}(P(\boldsymbol w))$ vs. $\mathrm{PPX}(P(\boldsymbol w, \boldsymbol G))$). Weights of LTLM and STLM when interpolated with MKN LM are shown on Figure \ref{fig:weights}.

\begin{table*}[Ht!]
\centering
\setlength{\tabcolsep}{0.71em}
\small
\begin{tabular}{llllllllll}
\hline
 \bf everything &             \bf has &          \bf  beauty &             \bf   , &         \bf   but &             \bf   not &     \bf   everyone &         \bf   sees &            \bf    it &        \bf    .    \\
\hline
              it &                's &              one &               , &               but &               was &                he &               saw &               him &                  .  \\
            that &                 is &             thing &                 ; &            course &                it &                 i &               made &                it &                  !  \\
             let &                was &              life &              -- &            though &               not &                she &              found &                her &               ...  \\
           there &            knows &               name &                 - &                or &              this &               they &               took &               them &                '  \\
       something &            really &             father &               ... &           perhaps &             that &              that &            gave &             his &               what  \\
         nothing &              says &             mother &                 : &                and &              the &                it &              told &               me &                 `` \\
      everything &              comes &               way &                 – &             maybe &               now &              who &               felt &                 a &               how  \\
            here &             does &              wife &                  ( &           although &              had &              you &           thought &                out &                why  \\
         someone &             gets &             place &                  ? &                yet &             $<$unk$>$ &            someone &               knew &              that &                 --  \\
             god &               has &              idea &             naught &             except &               all &              which &              heard &           himself &                 -  \\
\hline
\end{tabular}
\caption{\label{tab:substitutions}Ten most probable word substitutions on each position in the sentence "\emph{Everything has beauty, but not everyone sees it.}" produced by 1000 roles LTLM with the deterministic inference.}
\end{table*}

From the tables we can see several important findings. Standalone LTLM performs worse than MKN on both languages, however their combination leads to dramatic improvements compared with other LMs. Best results are achieved by 4-gram MKN interpolated with 1000 roles LTLM and the deterministic inference. The perplexity was improved by approximately $46\%$ on English and $49\%$ on Czech compared with standalone MKN. The deterministic inference outperformed the non-deterministic one in all cases. LTLM also significantly outperformed STLM where the syntactic dependency trees were provided as a prior knowledge. The joint learning of syntax and semantics of a sentence proved to be more suitable for predicting the words.

An in-depth analysis of semantic and syntactic properties of LTLM is beyond the scope of this paper. For better insight into the behavior of LTLM, we show the most probable word substitutions for one selected sentence (see Table \ref{tab:substitutions}). We can see that the original words are often on the front positions. Also it seems that LTLM is more syntactically oriented, which confirms claims from \cite{Levy:2014,Pado:2007}, but to draw such conclusions a deeper analysis is required.
The properties of the model strongly depends on the number of distinct roles. We experimented with maximally 1000 roles. To catch the meaning of various words in natural language, more roles may be needed. However, with our current implementation, it was intractable to train LTLM with more roles in a reasonable time. Training 1000 roles LTLM took up to two weeks on a powerful computational unit.

\section{Conclusion and Future Work\label{sec:conclusion}}

In this paper we introduced the Latent Tree Language Model. Our model discovers the latent tree structures hidden in natural text and uses them to predict the words in a sentence. Our experiments with English and Czech corpora showed dramatic improvements in the predictive ability compared with standalone Modified Kneser-Ney LM. Our Java implementation is available for research purposes at \url{https://github.com/brychcin/LTLM}.

It was beyond the scope of this paper to explicitly test the semantic and syntactic properties of the model. As the main direction for future work we plan to investigate these properties for example by comparison with human-assigned judgments. Also, we want to test our model in different NLP tasks (e.g. speech recognition, machine translation, etc.). 

We think that the role-by-role distribution should depend on the distance between the parent and the child, but our preliminary experiments were not met with success. We plan to elaborate on this assumption. Another idea we want to explore is to use different distributions as a prior to multinomials. For example, \newcite{Blei:2006} showed that the logistic-normal distribution works well for topic modeling because it captures the correlations between topics. The same idea might work for roles.

\section*{Acknowledgments}

This publication was supported by the project LO1506 of the Czech Ministry of Education, Youth and Sports. Computational resources were provided by the CESNET LM2015042 and the CERIT Scientific Cloud LM2015085, provided under the programme "Projects of Large Research, Development, and Innovations Infrastructures". Lastly, we would like to thank the anonymous reviewers for their insightful feedback.

\bibliography{emnlp2016}

\begin{thebibliography}{}

\bibitem[\protect\citename{Bengio \bgroup et al.\egroup }2003]{Bengio:2003}
Yoshua Bengio, R{\'e}jean Ducharme, Pascal Vincent, and Christian Janvin.
\newblock 2003.
\newblock A neural probabilistic language model.
\newblock {\em Journal of Machine Learning Research}, 3:1137--1155, March.

\bibitem[\protect\citename{Blei and Lafferty}2006]{Blei:2006}
David~M. Blei and John~D. Lafferty.
\newblock 2006.
\newblock Correlated topic models.
\newblock In {\em In Proceedings of the 23rd International Conference on
  Machine Learning}, pages 113--120. MIT Press.

\bibitem[\protect\citename{Bojar \bgroup et al.\egroup }2012]{czeng10:lrec2012}
Ond{\v{r}}ej Bojar, Zden{\v{e}}k {\v{Z}}abokrtsk{\'{y}}, Ond{\v{r}}ej
  Du{\v{s}}ek, Petra Galu{\v{s}}{\v{c}}{\'{a}}kov{\'{a}}, Martin Majli{\v{s}},
  David Mare{\v{c}}ek, Ji{\v{r}}{\'{\i}} Mar{\v{s}}{\'{\i}}k, Michal
  Nov{\'{a}}k, Martin Popel, and Ale{\v{s}} Tamchyna.
\newblock 2012.
\newblock The joy of parallelism with czeng 1.0.
\newblock In {\em Proceedings of the Eight International Conference on Language
  Resources and Evaluation (LREC'12)}, Istanbul, Turkey, may. European Language
  Resources Association (ELRA).

\bibitem[\protect\citename{Brown \bgroup et al.\egroup }1992]{Brown:1992}
Peter~F. Brown, Peter~V. deSouza, Robert~L. Mercer, Vincent J.~Della Pietra,
  and Jenifer~C. Lai.
\newblock 1992.
\newblock Class-based n-gram models of natural language.
\newblock {\em Computational Linguistics}, 18:467--479.

\bibitem[\protect\citename{Brychc\'{i}n and Konop\'{i}k}2014]{Brychcin:2014}
Tom\'a\v{s} Brychc\'{i}n and Miloslav Konop\'{i}k.
\newblock 2014.
\newblock Semantic spaces for improving language modeling.
\newblock {\em Computer Speech \& Language}, 28(1):192--209.

\bibitem[\protect\citename{Brychc\'{i}n and Konop\'{i}k}2015]{brychcin:2015}
Tom\'a\v{s} Brychc\'{i}n and Miloslav Konop\'{i}k.
\newblock 2015.
\newblock Latent semantics in language models.
\newblock {\em Computer Speech \& Language}, 33(1):88--108.

\bibitem[\protect\citename{Chen and Goodman}1998]{Chen:1998}
Stanley~F. Chen and Joshua~T. Goodman.
\newblock 1998.
\newblock An empirical study of smoothing techniques for language modeling.
\newblock Technical report, Computer Science Group, Harvard University.

\bibitem[\protect\citename{Cohen \bgroup et al.\egroup }2009]{Cohen:2009}
Shay~B. Cohen, Kevin Gimpel, and Noah~A. Smith.
\newblock 2009.
\newblock Logistic normal priors for unsupervised probabilistic grammar
  induction.
\newblock In {\em Advances in Neural Information Processing Systems 21}, pages
  1--8.

\bibitem[\protect\citename{Dempster \bgroup et al.\egroup }1977]{Dem:1977}
Arthur~P. Dempster, N.~M. Laird, and D.~B. Rubin.
\newblock 1977.
\newblock Maximum likelihood from incomplete data via the em algorithm.
\newblock {\em Journal of the Royal Statistical Society. Series B},
  39(1):1--38.

\bibitem[\protect\citename{Deschacht \bgroup et al.\egroup
  }2012]{Deschacht2012384}
Koen Deschacht, Jan~De Belder, and Marie-Francine Moens.
\newblock 2012.
\newblock The latent words language model.
\newblock {\em Computer Speech \& Language}, 26(5):384--409.

\bibitem[\protect\citename{Harris}1954]{Harris:1954}
Zellig Harris.
\newblock 1954.
\newblock Distributional structure.
\newblock {\em Word}, 10(23):146--162.

\bibitem[\protect\citename{Headden~III \bgroup et al.\egroup
  }2009]{Headden:2009}
William~P. Headden~III, Mark Johnson, and David McClosky.
\newblock 2009.
\newblock Improving unsupervised dependency parsing with richer contexts and
  smoothing.
\newblock In {\em Proceedings of Human Language Technologies: The 2009 Annual
  Conference of the North American Chapter of the Association for Computational
  Linguistics}, pages 101--109, Boulder, Colorado, June. Association for
  Computational Linguistics.

\bibitem[\protect\citename{K\"ubler \bgroup et al.\egroup }2009]{kubler09}
Sandra K\"ubler, Ryan McDonald, and Joakim Nivre.
\newblock 2009.
\newblock Dependency parsing.
\newblock {\em Synthesis Lectures on Human Language Technologies}, 2(1):1--127.

\bibitem[\protect\citename{Lange and Lei{\ss}}2009]{Lange09}
Martin Lange and Hans Lei{\ss}.
\newblock 2009.
\newblock To cnf or not to cnf? an efficient yet presentable version of the cyk
  algorithm.
\newblock {\em Informatica Didactica}, 8.

\bibitem[\protect\citename{Levy and Goldberg}2014]{Levy:2014}
Omer Levy and Yoav Goldberg.
\newblock 2014.
\newblock Dependency-based word embeddings.
\newblock In {\em Proceedings of the 52nd Annual Meeting of the Association for
  Computational Linguistics (Volume 2: Short Papers)}, pages 302--308,
  Baltimore, Maryland, June. Association for Computational Linguistics.

\bibitem[\protect\citename{Mare\v{c}ek and Straka}2013]{Marecek:2013}
David Mare\v{c}ek and Milan Straka.
\newblock 2013.
\newblock Stop-probability estimates computed on a large corpus improve
  unsupervised dependency parsing.
\newblock In {\em Proceedings of the 51st Annual Meeting of the Association for
  Computational Linguistics (Volume 1: Long Papers)}, pages 281--290, Sofia,
  Bulgaria, August. Association for Computational Linguistics.

\bibitem[\protect\citename{Martin \bgroup et al.\egroup }1998]{Martin:1998}
Sven Martin, Jorg Liermann, and Hermann Ney.
\newblock 1998.
\newblock Algorithms for bigram and trigram word clustering.
\newblock {\em Speech Communication}, 24(1):19--37.

\bibitem[\protect\citename{McDonald \bgroup et al.\egroup }2005]{McDonald:2005}
Ryan McDonald, Fernando Pereira, Kiril Ribarov, and Jan Haji\v{c}.
\newblock 2005.
\newblock Non-projective dependency parsing using spanning tree algorithms.
\newblock In {\em Proceedings of the Conference on Human Language Technology
  and Empirical Methods in Natural Language Processing}, HLT '05, pages
  523--530, Stroudsburg, PA, USA. Association for Computational Linguistics.

\bibitem[\protect\citename{Mikolov \bgroup et al.\egroup }2010]{Mikolov:2010}
Tom\'{a}\v{s} Mikolov, Martin Karafi\'{a}t, Luk\'{a}\v{s} Burget, Jan
  \v{C}ernock\'{y}, and Sanjeev Khudanpur.
\newblock 2010.
\newblock Recurrent neural network based language model.
\newblock In {\em Proceedings of the 11th Annual Conference of the
  International Speech Communication Association (INTERSPEECH 2010)}, volume
  2010, pages 1045--1048. International Speech Communication Association.

\bibitem[\protect\citename{Mikolov \bgroup et al.\egroup }2011]{Mikolov:2011}
Tom\'{a}\v{s} Mikolov, Stefan Kombrink, Luk\'{a}\v{s} Burget, Jan
  \v{C}ernock{\'{y}}, and Sanjeev Khudanpur.
\newblock 2011.
\newblock Extensions of recurrent neural network language model.
\newblock In {\em Proceedings of the {IEEE} International Conference on
  Acoustics, Speech, and Signal Processing}, pages 5528--5531, Prague Congress
  Center, Prague, Czech Republic.

\bibitem[\protect\citename{Minka}2003]{Minka03}
Thomas~P. Minka.
\newblock 2003.
\newblock Estimating a dirichlet distribution.
\newblock Technical report.

\bibitem[\protect\citename{Pad\'{o} and Lapata}2007]{Pado:2007}
Sebastian Pad\'{o} and Mirella Lapata.
\newblock 2007.
\newblock Dependency-based construction of semantic space models.
\newblock {\em Computational Linguistics}, 33(2):161--199, June.

\bibitem[\protect\citename{Popel and Mare\v{c}ek}2010]{Popel:2010}
Martin Popel and David Mare\v{c}ek.
\newblock 2010.
\newblock Perplexity of n-gram and dependency language models.
\newblock In {\em Proceedings of the 13th International Conference on Text,
  Speech and Dialogue}, TSD'10, pages 173--180, Berlin, Heidelberg.
  Springer-Verlag.

\bibitem[\protect\citename{Spitkovsky \bgroup et al.\egroup
  }2010]{Spitkovsky:2010}
Valentin~I. Spitkovsky, Hiyan Alshawi, Daniel Jurafsky, and Christopher~D.
  Manning.
\newblock 2010.
\newblock Viterbi training improves unsupervised dependency parsing.
\newblock In {\em Proceedings of the Fourteenth Conference on Computational
  Natural Language Learning}, pages 9--17, Uppsala, Sweden, July. Association
  for Computational Linguistics.

\bibitem[\protect\citename{Spitkovsky \bgroup et al.\egroup
  }2011]{Spitkovsky:2011}
Valentin~I. Spitkovsky, Hiyan Alshawi, Angel~X. Chang, and Daniel Jurafsky.
\newblock 2011.
\newblock Unsupervised dependency parsing without gold part-of-speech tags.
\newblock In {\em Proceedings of the 2011 Conference on Empirical Methods in
  Natural Language Processing}, pages 1281--1290, Edinburgh, Scotland, UK.,
  July. Association for Computational Linguistics.

\bibitem[\protect\citename{Whittaker and Woodland}2003]{Whittaker:2003}
Edward W.~D. Whittaker and Philip~C. Woodland.
\newblock 2003.
\newblock Language modelling for russian and english using words and classes.
\newblock {\em Computer Speech \& Language}, 17(1):87--104.

\end{thebibliography}
\bibliographystyle{emnlp2016}

\end{document}